\documentclass[runningheads]{llncs}
\usepackage[T1]{fontenc}
\usepackage{graphicx}
\usepackage{algorithm}
\usepackage{csquotes}
\usepackage{algpseudocode}
\usepackage{amssymb}
\usepackage{amsmath}
\usepackage{color}
\usepackage{booktabs}
\usepackage{comment}
\usepackage{tabularx}

\begin{document}
\title{Food Image Segmentation with LLM-Derived Ingredient Labels and Multimodal Fusion}
%
%
\author{Jui-Feng Chi \and
Wei-Ta Chu \orcidID{0000-0001-5722-7239} \and 
Sheng-Long Lin}
\authorrunning{Chi et al.}
%

\institute{National Cheng Kung University, Tainan, Taiwan \\
\email{aj714900@gmail.com}, 
\email{wtchu@gs.ncku.edu.tw}, 
\email{opeddie22@gmail.com}
}

\maketitle              
\begin{abstract}
Food image segmentation plays a vital role in health-related applications such as nutrition tracking and personalized health monitoring. However, existing models often underperform on visually similar ingredients and rare food categories. To address this issue, we propose two plug-and-play multimodal modules that enhance the segmentation performance by leveraging ingredient labels inferred from food images using large language models (LLMs). The first module, called LIM-F (Language Injection Module for Features), is designed to pair with any image encoder that produces multi-layer outputs (e.g., Swin Transformer), while the second module, LIM-Q (Language Injection Module for Queries), targets Mask2Former-style Transformer-based decoders. Both modules enable training without the need for pre-aligning images with text by directly injecting semantic ingredient information into the visual analysis pipeline. On the FoodSeg103 benchmark, the proposed method achieves state-of-the-art performance. Specifically, integrating LIM-Q into the Mask2Former decoder with a Swin-L image encoder yields a mean Intersection over Union (mIoU) of 55.0. LIM-F also demonstrates strong generalization and competitive performance, reaching an mIoU of 54.4 under the same model (Swin-L+Mask2Former). Furthermore, its applicability extends beyond Transformer-based decoders, as evidenced by an improvement from 47.7 to 49.8 mIoU when integrated into a CNN-based architecture. Notably, the improved segmentation accuracy is achieved with only a moderate (at most 3.8 GB) increase in the GPU memory consumption during training. Thus, the proposed approach offers a practical and scalable solution for fine-grained food understanding.

\end{abstract}
\section{Introduction}
\label{sec:intro}
Food image segmentation is a fundamental task in computer vision with growing significance in healthcare, dietary assessment, and nutrition tracking. Accurately segmenting food items enables applications like calorie estimation, dietary monitoring, and intelligent meal logging. However, food segmentation poses unique challenges owing to the high inter-class similarity and intra-class variability of many ingredients. For example, tofu and cheese are difficult to distinguish by using only appearance-based features.

Recent studies have utilized vision-language models to improve the segmentation accuracy. For instance, OVFoodSeg \cite{wu24} enhances open-vocabulary segmentation by using image-aware textual representations. However, aligned image-text pairs are required to train the model. IngredSAM \cite{chen24} takes pre-defined ingredient images as the reference to generate initial prompts, and relies on the Segmenting Anything Model (SAM) \cite{kirillov23} to achieve food segmentation. Appropriate ingredient images should be prepared in advance. ReLeM \cite{wu21}, introduced as part of the FoodSeg103 benchmark, injects ingredient-level semantics into the visual pipeline using recipe-image pairs from the Recipe1M+ dataset \cite{marin21}. It also relies on extensive multimodal pre-training and curated datasets. 

These methods represent important steps toward the integration of text semantic information into food image segmentation pipelines. However, they rely on large-scale paired datasets such as Recipe1M+, involve intricate training routines, or are often tied to specific backbone architectures. These factors limit their scalability and adaptability, particularly in domains that lack richly aligned corpora.

To address these limitations, we propose two plug-and-play multimodal fusion modules that introduce ingredient information into the segmentation pipeline. In this context, plug-and-play refers to the ability to integrate the modules into existing segmentation architectures without modifying the backbone network or annotation format. This modular design allows for seamless incorporation into standard training and inference workflows, thereby reducing the implementation complexity and enhancing the practical applicability. The first module, Language Injection Module for Features (LIM-F), conducts text-guided cross-attention by fusing ingredient label information with intermediate visual features. Its design is compatible with general segmentation models like K-Net + UPerNet \cite{zhang21}\cite{xiao18}. The second module, LIM-Q (Language Injection Module for Queries), injects language-derived features to enhance query vectors in a Transformer-based decoder to achieve better segmentation. 

Both modules use ingredient class labels generated by a large language model (ChatGPT in the present case), thereby eliminating the need for the manual alignment of image-text pairs or recipe datasets \cite{wu24}\cite{marin21}. The proposed design is light-weight, incurs only a moderate (at most 3.8 GB) increase in the GPU memory overhead, and converges faster with fewer training iterations. 

Three integration variants are evaluated in the experiments: LIM-F integrated with either K-Net \cite{zhang21}\cite{xiao18} or Mask2Former \cite{cheng22}, and LIM-Q integrated with Mask2Former, all using a Swin-L backbone \cite{liu21}. We show that all variants outperform strong baselines on the FoodSeg103 benchmark. LIM-Q achieves a new state-of-the-art mIoU of 55.0, surpassing prior methods such as BEiT v2 Large (49.4) \cite{sinha23} and Swin-TUNA (50.6) \cite{chen25}, while LIM-F closely follows with a mIoU of 54.4. 


\section{Related Works}
\label{sec:related}
\subsection{Semantic Segmentation Models} 
Early semantic segmentation models were predominantly built on convolutional neural networks (CNNs), such as PSPNet \cite{zhao17} and DeepLab \cite{chen17}. Among CNN-based methods, K-Net \cite{zhang21} is notable for its dynamic kernel updating mechanism, which supports both instance and panoptic segmentation. More recently, the emergence of Transformer-based architectures has led to significant performance improvements across segmentation tasks. In particular, Mask2Former \cite{cheng22} has demonstrated state-of-the-art results in panoptic, instance, and semantic segmentation by leveraging masked attention mechanisms and learnable query tokens. 

\subsection{Transformers for Vision and Multimodal Tasks}
The Transformer architecture, driven by self-attention, has become the de facto backbone in both natural language processing and computer vision. Vision-specific variants, such as ViT \cite{dos21} and Swin Transformer \cite{liu21}, have shown that  Transformers can effectively model both image patches and multi-scale visual features. These developments have laid the groundwork for cross-attention mechanisms in multimodal systems, where queries from one modality (e.g., image tokens) attend to keys and values from another (e.g., text tokens). This approach enables fine-grained alignment between different modalities, as demonstrated in models such as ViLBERT \cite{lu19} and UNITER \cite{chen20}. In dense prediction tasks, Mask2Former \cite{cheng22} and related architectures further extend this paradigm by incorporating cross-attention between visual features and learnable query tokens, thereby enabling more precise segmentation. 

\subsection{Vision Systems Powered by Large Language Models}
Recent studies have shown that integrating LLMs into vision systems is both feasible and effective for enhancing visual understanding. For example, MiniGPT-4 \cite{zhu23} and LLaVA \cite{liu23} align vision encoders (e.g., CLIP \cite{radford21} or ViT \cite{dos21}) with LLMs such as Vicuna \cite{vicuna2023} to support region-level captioning and visual question answering. KOSMOS-2 \cite{peng23} grounds entities in visual inputs using a multimodal GPT-style decoder, while GPT-4 \cite{gpt23} demonstrates strong zero-shot reasoning in vision-language tasks through large-scale decoder-only modeling. These models highlight the growing potential of LLMs to enrich visual representations through the provision of semantic context. Inspired by this line of research, our study integrates a BERT-based text encoder \cite{devlin19} (for its simplicity) with a visual backbone, enabling ingredient-level guidance via cross-modal attention, thereby avoiding the need for paired image-text pretraining. 

\subsection{Text-Guided Food Segmentation Models} 
Despite growing interest in multimodal learning, few studies have explored using textual cues in food segmentation tasks. One prominent example is ReLeM \cite{wu21}, which applies contrastive training on paired recipe narratives and dish images from the Recipe1M+ dataset. By embedding ingredient and cooking method semantics into the visual backbone, ReLeM produces an ingredient-aware encoder that can be paired with various segmentation networks. Another example is OVFoodSeg \cite{wu24}, which is an open-vocabulary approach that employs a transformer-based module to convert image encoder outputs into a representation compatible with the text encoder. This design enables the ingredient tokens to dynamically adapt to the visual context. 

Although both methods demonstrate the value of ingredient semantics in food segmentation, they rely on complex pipelines and curated paired datasets. In contrast, our proposed framework achieves similar cross-modal fusion through a lightweight, plug-and-play design that removes the dependency on large-scale vision-language pretraining and paired datasets. 


\section{Method}
\label{sec:method}
Our proposed method integrates the ingredient-level textual features generated by an LLM into a semantic segmentation architecture through two complementary modules: LIM-F, a general-purpose module applicable to any segmentation model with multi-level output backbones, and LIM-Q, a decoder-specific design tailored for a Transformer-based decoder. Both modules are designed to be lightweight and architecture-compatible, enabling efficient multimodal learning without special pretraining or external image-text alignment.

\begin{figure}[t]
   \centering
   \includegraphics[width=0.98\linewidth]{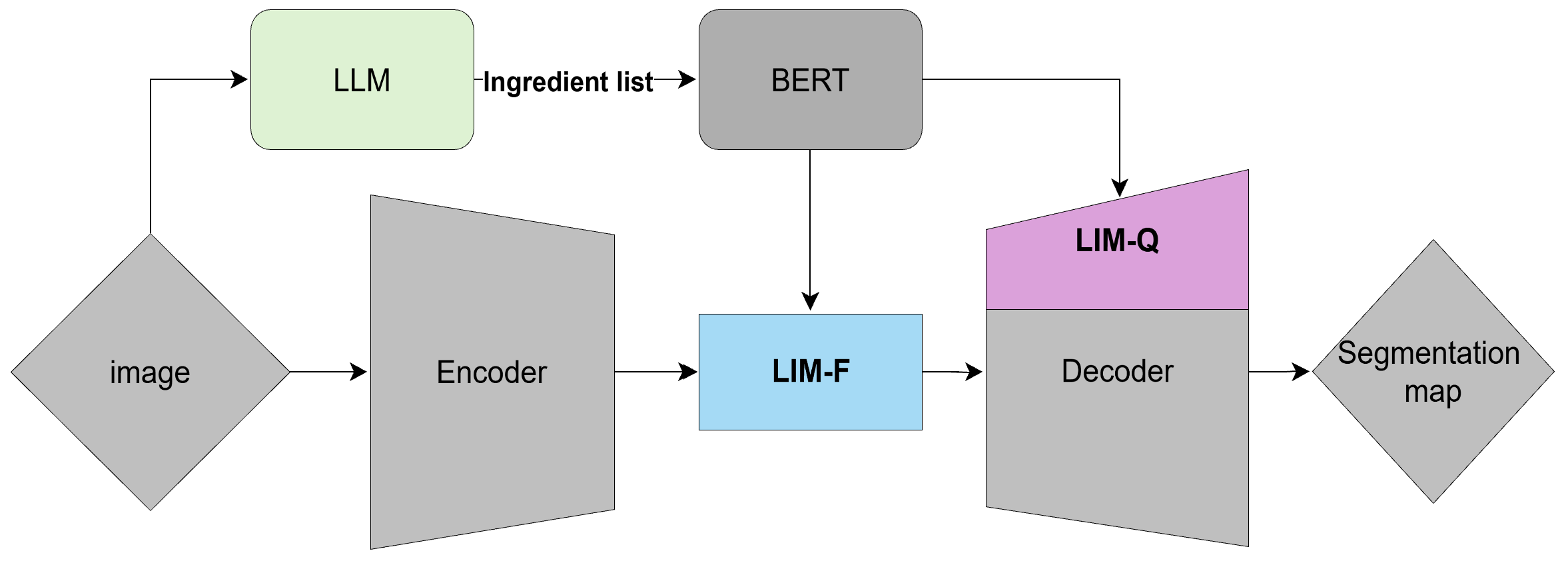}
   \caption{Overview of the proposed multimodal segmentation framework. An input image is sent to both a vision encoder and an LLM, which generates ingredient labels that are then encoded by BERT. The language features are injected into the segmentation pipeline via either LIM-F or LIM-Q. LIM-F fuses feature maps after the encoder, while LIM-Q embeds language information into the decoder. Both the encoder and decoder are modular and can be flexibly replaced.}
   \label{fig:overview}
\end{figure}

\subsection{Overview}
Figure~\ref{fig:overview} shows an overview of the proposed system, which consists of a dual-branch structure composed of a visual encoder and a language encoder. The visual branch uses Swin-L \cite{liu21} pretrained on the ImageNet-22k dataset and then finetuned (any other backbone with multi-level outputs can also be used) as the image encoder to produce multi-scale visual feature maps. Meanwhile, the language branch takes a list of ingredient classes generated by an online LLM (GPT-o4 mini in this study) as the input and encodes these class labels using a BERT model \cite{devlin19} to produce word embeddings\footnote{The BERT-large-uncased model downloaded from Hugging Face, https://huggingface.co/google-bert/bert-large-uncased}. These embeddings serve as the source of semantic guidance for the fusion modules. 

Depending on the backbone and decoder design, one of the two language injection modules is used, namely LIM-F, which introduces language supervision into intermediate visual features for models with multi-level output backbones, or LIM-Q, which modulates Transformer decoder queries using language features. Unlike LIM-F, LIM-Q operates only at the decoder level and is compatible with a wide range of backbone architectures. 

\begin{figure}[t]
   \centering
   \includegraphics[width=0.98\linewidth]{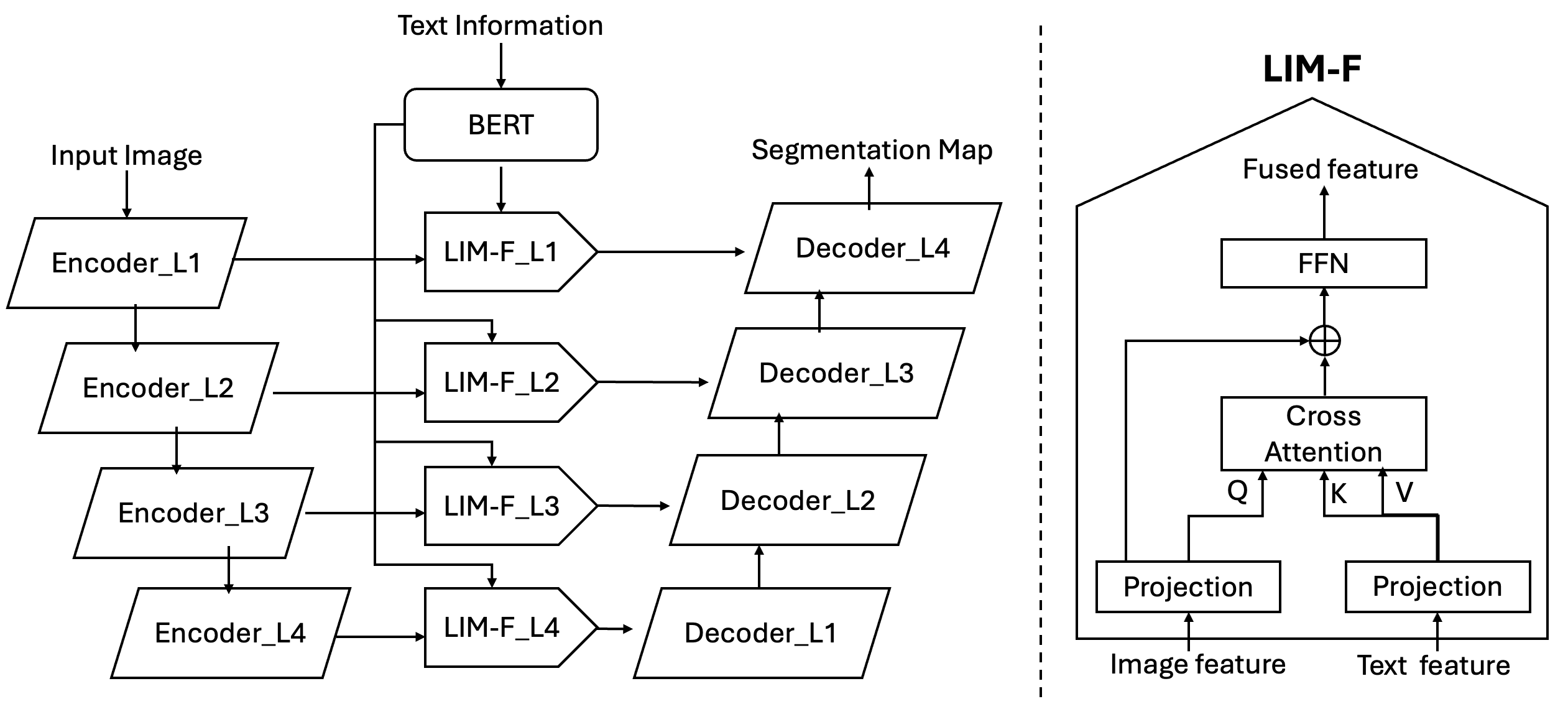}
   \caption{The segmentation framework with the LIM-F module. Feature maps from different levels are individually passed through a dedicated LIM-F block. These blocks apply cross-attention between the image features and BERT-encoded ingredient embeddings, allowing multi-scale semantic fusion before passing to the decoder. All query, key, and value vectors are layer-normalized to ensure stable feature distributions.}
   \label{fig:limf}
\end{figure}

\subsection{LIM-F: Feature-level Injection for Multi-Level Output Backbones}
Figure~\ref{fig:limf} illustrates the framework with the LIM-F module, which is designed to introduce ingredient-level text semantics into architectures with a multi-level output image encoder. Specifically, the Swin-L encoder generates four output levels (levels 0 to 3), each corresponding to a different spatial resolution and receptive field. At each level, an independent LIM-F module calculates cross-attention between visual and text information, where the visual features serve as queries and the BERT-encoded ingredient embeddings serve as keys and values. 

Dimensional alignment between the BERT embeddings and the visual feature map at each level is ensured using a linear projection layer. In general, the textual feature dimension needs to match that of the corresponding image feature map; therefore, the image feature map retains the same dimensionality before and after its own linear projection. The output of each cross-attention operation is added back to the visual stream via a residual connection, followed by a feed-forward network (FFN), and then passed to the corresponding decoder layer. The application of cross-attention at all levels enables multi-level features to attend to textual ingredient descriptions, incorporating the corresponding semantic cues into relevant pixels. 

Since LIM-F relies solely on the image encoder and imposes no constraints on the decoder, it offers broad adaptability across segmentation architectures. To demonstrate this property, we implement two configurations of LIM-F in our experiments. The first variant pairs Swin-L with K-Net + UPerNet \cite{zhang21}\cite{xiao18} as the decoder. Here, UPerNet serves as a conventional CNN-based decoder, while K-Net extends it with dynamic kernel updates, showcasing LIM-F's compatibility with traditional non-Transformer decoders. The second variant adopts Mask2Former, a Transformer-based decoder known for strong performance, to further examine LIM-F's adaptability and effectiveness across diverse decoder designs.

\begin{figure}[t]
   \centering
   \includegraphics[width=0.98\linewidth]{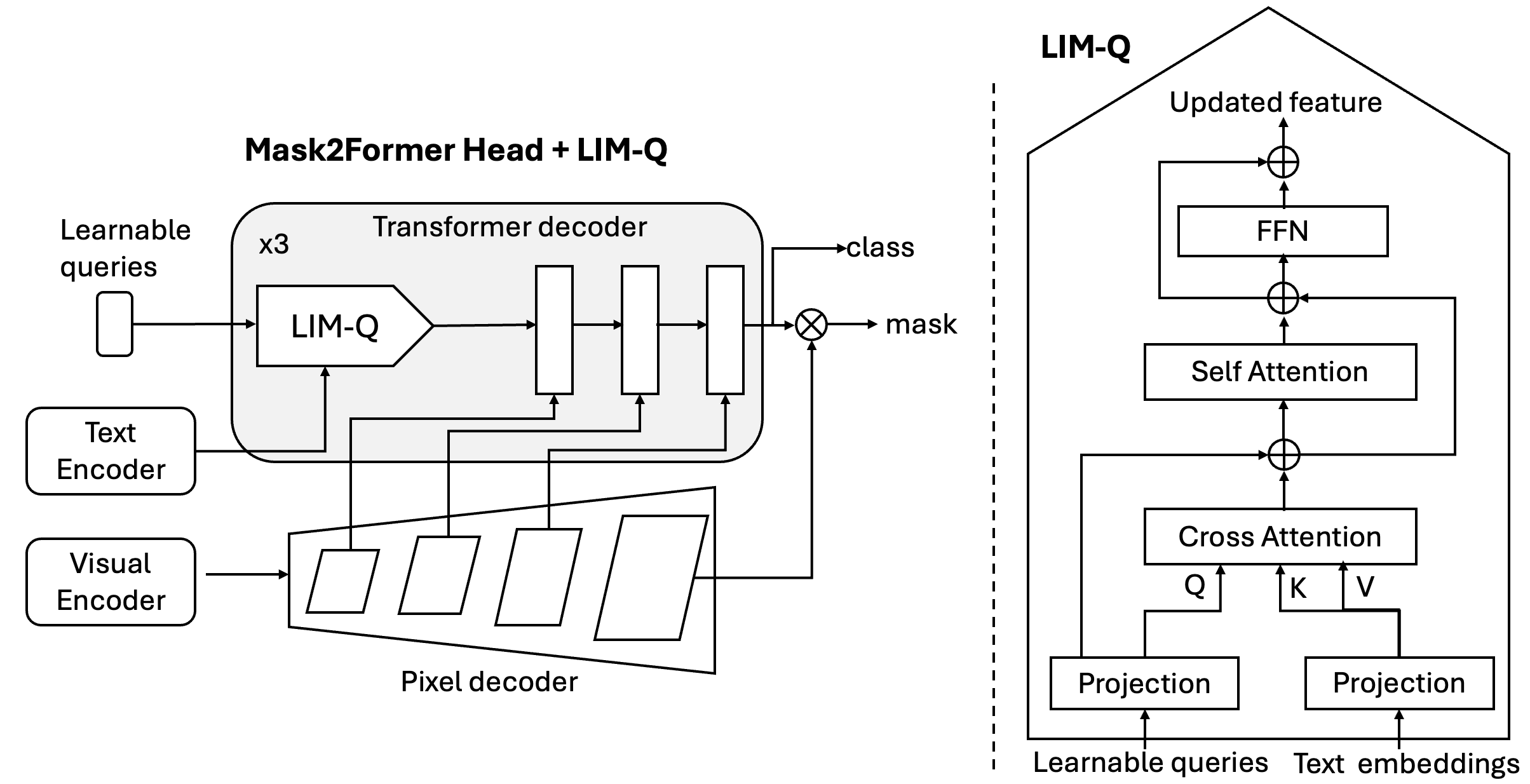}
   \caption{Each LIM-Q block is inserted into every decoding round of Mask2Former to enhance its ability to incorporate textual semantics. It includes a cross-attention to BERT-encoded ingredient embeddings, followed by self-attention and a feed-forward network (FFN). All Q, K, and V inputs are layer-normalized before entering each attention block.}
   \label{fig:limq}
\end{figure}
 
\subsection{LIM-Q: Query-level Injection for Transformer Decoders}
Figure~\ref{fig:limq} shows the framework with the LIM-Q module, which extends the conventional Mask2Former decoder structure. In Mask2Former, a set of learnable query tokens is updated over three rounds, with each round comprising three Transformer decoder layers. During each round, the query tokens attend to the feature maps produced at different levels by the pixel decoder in a round-robin manner, resulting in a total of nine layers. 

To incorporate guidance from ingredient labels, we introduce semantics-based guidance at the beginning of every round. Before the learnable queries attend to the three image feature levels, they first pass through a dedicated LIM-Q block. Each LIM-Q block first projects the learnable queries and BERT-encoded ingredient embeddings (as key and values) into a common space, where the feature dimensions are matched to the queries. The block then performs cross-attention to inject ingredient semantics into the queries, followed by a self-attention layer to model relations among the updated queries and a feed-forward network (FFN) for further refinement, with all stages wrapped in residual connections. 

The insertion of LIM-Q blocks in a round-robin manner ensures that semantic guidance from the LLM-generated text labels is consistently reinforced throughout decoding, effectively reminding the queries of the ingredient category they should predict. This mechanism mitigates common failure cases, such as confusing visually similar food classes, and helps achieve better class-level alignment.




\subsection{Loss Function}
Based on the proposed LIM-F and LIM-Q modules, we develop three variants of food segmentation methods: (1) K-Net with UPerNet + LIM-F; (2) Mask2Former + LIM-F; and (3) Mask2Former + LIM-Q. All three variants use Swin-L as the image encoder. 

Since different decoders often require different loss functions, we describe them separately. For the first variant, the loss function is the standard cross-entropy loss, which is commonly used in conventional decoders and aligns with the purpose of evaluating LIM-F's broad compatibility. For the second and third variants, the Mask2Former adopts the mask, the Dice, and the classification losses to guide model training. Specifically, the overall loss function is: 
\begin{equation}
\mathcal{L} = \lambda_{CE} \mathcal{L}_{CE} + \lambda_{dice} \mathcal{L}_{dice} + \lambda_{cls} \mathcal{L}_{cls},
\end{equation}
where $\mathcal{L}_{CE}$ is the pixel-wise Cross-Entropy loss for mask prediction, $\mathcal{L}_{dice}$ measures region-level overlap between predicted and ground-truth masks, and $\mathcal{L}_{cls}$ is the classification loss applied to the decoder's query-level category prediction. The weights $\lambda_{CE}$, $\lambda_{dice}$, and $\lambda_{cls}$ determine the relative contributions of each loss term. 

In the original configuration of Mask2Former, the weights $\lambda_{CE}$, $\lambda_{dice}$, and $\lambda_{cls}$ are set as 5, 5, and 2, respectively. When combining with the proposed LIM-F or LIM-Q modules, we found that appropriately increasing the weight $\lambda_{cls}$ gives more performance gain. This will be verified in the evaluation section. 


\section{Experiments}
\subsection{Implementation Details}
All variants were trained on the FoodSeg103 dataset \cite{wu21} with a batch size of 2 for 240,000 iterations. All experiments use Swin-L initialized with ImageNet-22k pretrained weights and BERT initialized with the pretrained BERT-large-uncased model from Hugging Face. For the pretrained Swin-L backbone and the last four layers of the pretrained BERT encoder (which were unfrozen for fine-tuning during the training process), a learning rate multiplier of 0.1 was applied. The parameters of the last four layers of the BERT encoder were trained without weight decay (decay set to 0.0), while the remaining BERT layers were kept frozen during training. All three variants use an AdamW optimizer.

Separately for different variants, the first variant uses a base learning rate of 6e-5, which was annealed to 1e-6 over the training schedule. For the second and third variants with Mask2Former, the training hyperparameters were assigned the same values as those in \cite{cheng22}, except for the loss weights. In addition, training was performed with an initial learning rate of 1e-4 decayed to 1e-6 using cosine annealing \cite{losh17} over the training schedule.



To provide language guidance, GPT-o4 mini\footnote{accessed in early May, 2025.} was used as the LLM to generate ingredient class labels. The input to the LLM consisted of a food image and the full list of 104 ingredient class names from FoodSeg103, along with a prompt asking the model to infer a few of the likely ingredients present in the food image and return them as a comma-separated list (e.g., \enquote{lettuce, potato, tomato}). This dynamic generation of image-specific textual description eliminates the need for curated image-text pairs and thus greatly reduces the training requirement. 


\begin{table}[t]
\centering
\caption{Performance comparison in terms of mIoU on the FoodSeg103 dataset.}
\begin{tabular}{l| l}
\hline
Models & mIoU \\
\hline
BEiTv2 Large \cite{sinha23} & 49.4\\
IngredSAM \cite{chen24} & 48.8 \\
FoodSAM \cite{lan23} & 46.4 \\
Swin-TUNA \cite{chen25} & 50.6 \\
\hline
K-Net (Swin-L) \cite{zhang21} & 47.7 \\
(1) K-Net (Swin-L) + LIM-F & 49.0 (+2.7\%) \\
\hline
Mask2Former (Swin-L) \cite{cheng22} & 51.9 \\
(2) Mask2Former (Swin-L) + LIM-F & 54.4 (+4.8 \%)\\
(3) Mask2Former (Swin-L) + LIM-Q & 55.0 (+6.0 \%) \\
\hline
\end{tabular}
\label{tb:performance}
\end{table}

\begin{figure}
   \centering
   \includegraphics[width=0.8\linewidth]{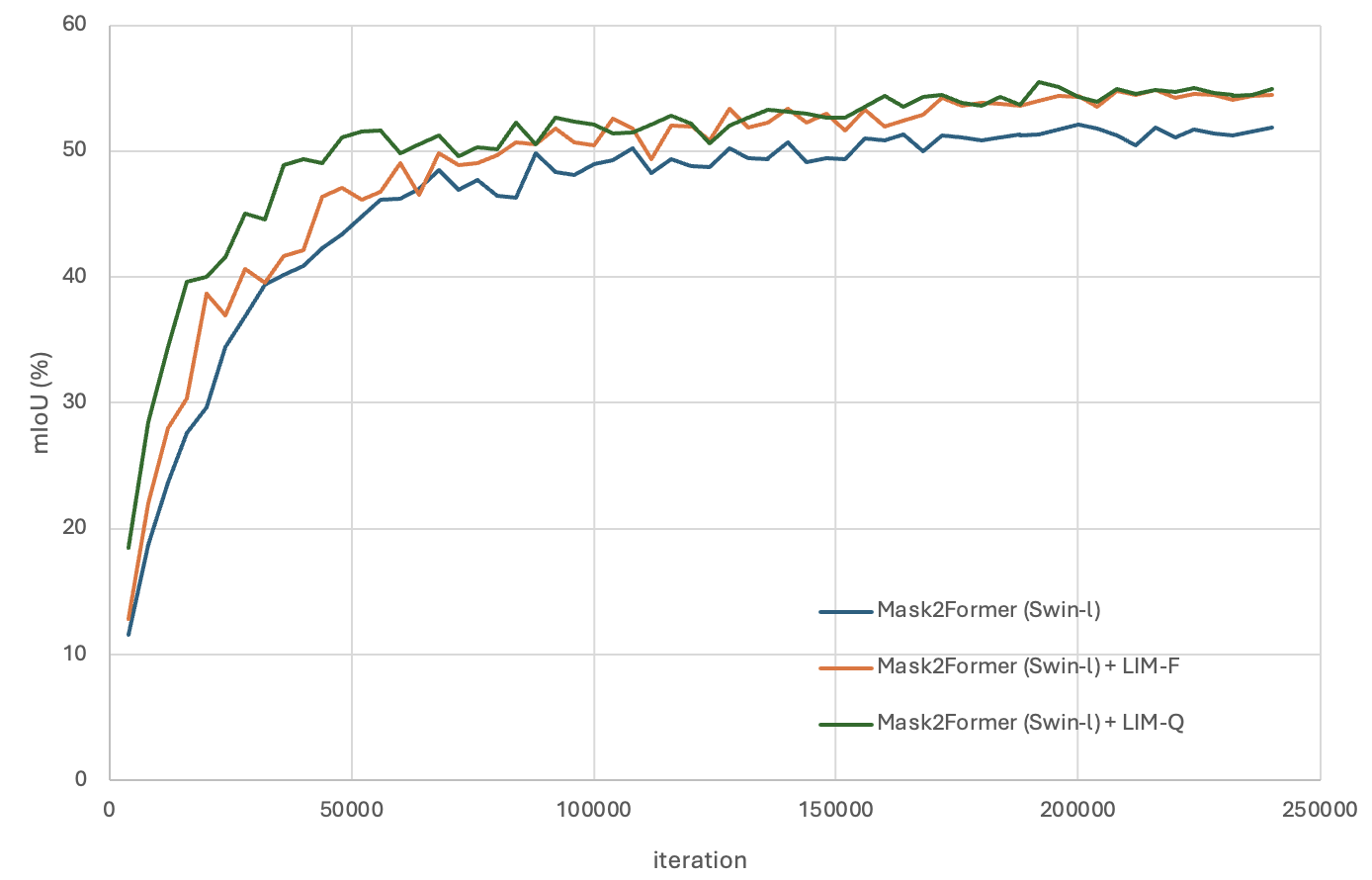}
   \caption{This figure compares the evolution of mIoUs over the training process, with the baseline Mask2Former, and its variants with LIM-F and LIM-Q modules. The two LIM-augmented models consistently outperform the baseline across training iterations, achieving higher final mIoU scores of 54.44 (LIM-F) and 54.96 (LIM-Q), compared to 51.86 for the original. }
   \label{fig:convergence}
\end{figure}

\subsection{Quantitative Results and Convergence Behavior}
We evaluate the segmentation performance in terms of mean Intersection over Union (mIoU) in Table~\ref{tb:performance}. As can be seen, the third variant (3) Mask2Former + LIM-Q achieves the highest mIoU of 55.0, surpassing strong baselines such as BEiTv2 Large (49.4), IngredSAM (48.8), Swin-TUNA (50.6), and FoodSAM (46.4) \cite{lan23}. LIM-F is also proven effective, achieving 49.0 when integrated with K-Net in the first variant and 54.4 when combined with Mask2Former in the second variant. 

Notably, although LIM-F and LIM-Q obtain similar mIoUs when integrated with Mask2Former in the second and third variants, LIM-Q demonstrates faster convergence during training, as shown in Figure~\ref{fig:convergence}. This can be attributed to the more aggressive text-guided information injection strategy in LIM-Q, in which the BERT-derived semantic cues are explicitly attended to by the learnable queries via cross-attention in the transformer decoder. On the other hand, LIM-F relies on gradually blending the textual semantics provided by the LLM into the multi-scale image features, allowing the query tokens to implicitly discover the semantic correspondences. This design difference allows LIM-Q to benefit from stronger and more direct language supervision in the early stages of training. 

Table~\ref{tb:ingredient} shows how LIM modules improve the segmentation performance for specific ingredients. To ensure a consistent baseline, the second and third variants are used for comparison in this study. We see that both LIM modules significantly improve the segmentation of rare or visually ambiguous ingredients, thereby confirming that ingredient-level language information helps compensate for weak or ambiguous visual signals.

\begin{table}[t]
\centering
\caption{LIM-F and LIM-Q greatly improve segmentation on rare ingredients shown in the table, each occupying less than 1.2\% of the training set, showing better recognition than the original Mask2Former.}
\begin{tabular}{l| l |l |l }
\hline
Sample ingredients  & Mask2Former & (2) Mask2Former + LIM-F & (3) Mask2Former + LIM-Q  \\
\hline
\hline 
date & 1.28 & 88.93 & 77.31  \\
melon & 7.95 & 60.90 & 52.27 \\
dried cranberries & 21.5 & 42.10 & 39.07 \\
pear & 31.08 & 58.04 & 80.57 \\
fig & 48.78 & 79.27 & 67.40  \\
\hline
\end{tabular}
\label{tb:ingredient}
\end{table}

The basic Mask2Former requires approximately 19 GB of GPU memory, while Mask2Former+LIM-F and Mask2Former+LIM-Q require 22.5 GB and 20 GB, respectively. Therefore, the add-on modules incur only a moderate increase in the GPU memory overhead during training. In our experiments, the proposed architecture achieves notable performance gains without any special pretraining, such as the text-image semantic alignment commonly employed in other multimodal models, making it more implementer-friendly.


\subsection{Ablation Study}
We investigate the weights of different losses on the Mask2Former-based segmentation performance in the second and third experiment variants. Table~\ref{tb:weights} shows the performance variations when different weights are employed on the classification loss, and whether the BERT model is fine-tuned. For LIM-F in the second variant, increasing the classification loss weight from 2 to 4 yields the highest mIoU of 54.4. For LIM-Q in the third variant, a (5, 5, 6) setting produces the best result of 55.0. This suggests that the classification head benefits from stronger supervision when it is enriched with language information. 

Overall, the ablation test results show the importance of careful tuning of loss weighting and textual representation strategies to maximize the segmentation accuracy for multimodal architectures.

\begin{table}
\centering
\caption{Performance variations when different weights are employed on the classification loss, and whether the BERT model is fine-tuned. }
\begin{tabular}{l| c |c |l }
\hline
Model & $(\lambda_{CE}, \lambda_{dice}, \lambda_{cls})$ & Unfreeze BERT & mIoU  \\
\hline
Mask2Former + LIM-F & (5, 5, 2) &  & 51.4  \\
Mask2Former + LIM-F & (5, 5, 4) &  & 52.1 \\
Mask2Former + LIM-F & (5, 5, 4) & $\surd$ & \textbf{54.4} \\
Mask2Former + LIM-F & (5, 5, 6) & $\surd$ & 53.8 \\
\hline
\hline
Mask2Former + LIM-Q & (5, 5, 4) &  & 51.4  \\
Mask2Former + LIM-Q & (5, 5, 6) &  & 52.8 \\
Mask2Former + LIM-Q & (5, 5, 6) & $\surd$ & \textbf{55.0} \\
Mask2Former + LIM-Q & (5, 5, 8) & & 52.2 \\
\hline
\end{tabular}
\label{tb:weights}
\end{table}

\subsection{Qualitative Analysis}
\begin{figure}
   \centering
   \includegraphics[width=0.98\linewidth]{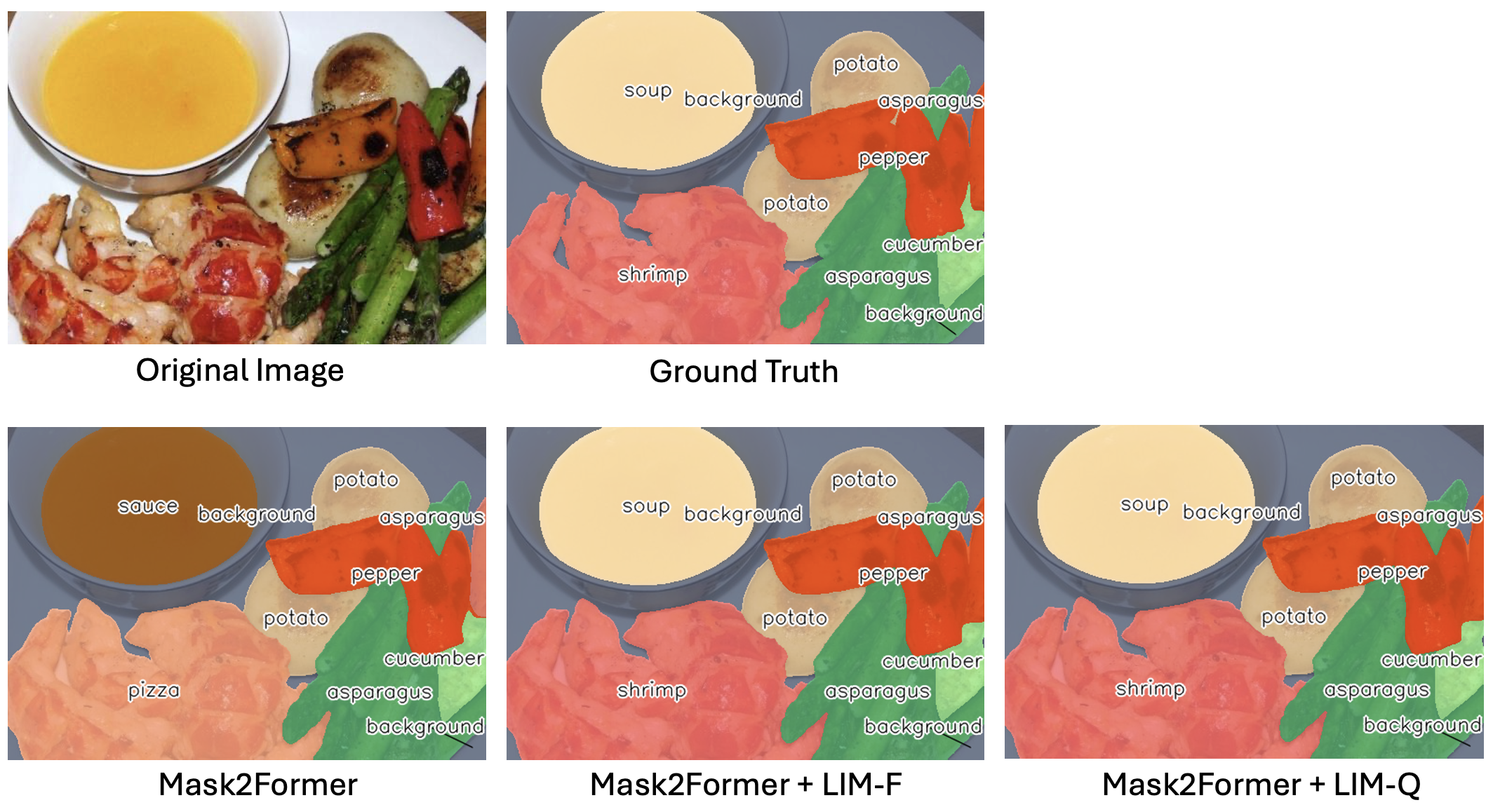}
   \caption{Sample segmentation results. }
   \label{fig:visualization}
\end{figure}

Figure~\ref{fig:visualization} shows sample segmentation results. We see that the Mask2Former model struggles to recognize the soup as sauce. With the proposed LIM modules, almost all regions are accurately segmented and recognized. In this example, the visual similarity between certain ingredient categories is particularly pronounced. The sauce in the upper-left bowl shares a similar color and texture with the soup, making it challenging for the baseline Mask2Former to distinguish them without additional semantic cues. Similarly, the grilled shrimp in the lower-left region exhibits a reddish-orange hue and segmented shape that closely resembles the appearance of pizza toppings under similar lighting conditions. These strong inter-class visual similarities contribute to the misclassifications observed in the baseline, while the incorporation of LIM-F and LIM-Q provides additional semantic constraints that help disambiguate these cases.  

\section{Conclusion}
\label{sec:conclusion}
This study introduces two lightweight, plug-and-play multimodal fusion modules, LIM-F and LIM-Q, that enhance food image segmentation through language guidance. LIM-F integrates linguistic semantics into the intermediate visual features produced by the image encoder, while LIM-Q injects language information directly into the transformer decoder queries. By leveraging ingredient lists generated by an LLM and encoded by a BERT model, along with the proposed LIM-F and LIM-Q modules, the proposed approach eliminates the need for costly image-text alignment and enables end-to-end training across a variety of backbone architectures. Experiments on the FoodSeg103 benchmark demonstrate that both modules clearly improve the segmentation accuracy, particularly for visually similar or rare ingredients. 

Overall, this study demonstrates the effectiveness of simple yet powerful language-injection strategies for semantic food segmentation. Future studies may explore adaptive token filtering, LLM-guided region proposals, or extensions to broader food-related tasks, such as volume estimation and cross-modal meal understanding. 

\textbf{Acknowledgement.}
This work was funded in part the National Science and Technology Council, Taiwan, under grants 114-2622-E-006-028, 114-2221-E-006-047-MY3, 114-2425-H-006-004, 114-2637-8-218-002, 113-2622-E-006-029, 113-2634-F-006-002, and 112-2221-E-006-136-MY3. 

%
%
%
\bibliographystyle{splncs04}
\bibliography{food}

\end{document}